\documentclass[runningheads]{llncs}

\usepackage{eccvabbrv}
\usepackage{eccv}

\usepackage{graphicx}
\usepackage{subcaption}
\usepackage{booktabs}

\usepackage[accsupp]{axessibility}  

\usepackage[pagebackref,breaklinks,colorlinks,citecolor=eccvblue]{hyperref}

\usepackage{orcidlink}

\begin{document}

\title{Human Action Recognition without Human} 

\titlerunning{Human Action Recognition without Human}

\author{Hirokatsu Kataoka\inst{1}\orcidlink{0000-0001-8844-165X} \and
Kensho Hara\inst{2,3}\orcidlink{0000-0001-6463-7738} \and
Yutaka Satoh\inst{3}\orcidlink{0000-0002-0638-0855}}

\authorrunning{H. Kataoka et al.}

\institute{National Institute of Advanced Industrial Science and Technology (AIST),\\Tsukuba, Ibaraki, Japan \\
\url{https://hirokatsukataoka.net/}}

\maketitle

\begin{abstract}
  This paper presents a novel evaluation of human action and its context that unveils a lack of current deep neural network-based action recognition. Particularly in human action recognition in videos, two-stream architecture from RGB (spatial stream) and optical flow (temporal stream) is becoming a representative technique  Although the two-stream ConvNets and improved approaches perform well with video datasets, the spatial RGB input seems to be too strong. For instance, the spatial stream in two-stream ConvNets achieved over 70\% on the UCF101 dataset. We believe that the problem of human action recognition in videos has been replaced by simple scene recognition without any temporal information. In this paper, we evaluate two different scenarios in human action recognition, (i) the effects of context when eliminating human areas , and (ii) the performance of pure human motion using recent motion representation techniques. We obtained the interesting experimental result that an only-context feature can achieve a performance of 65.48\% on UCF101 without a human.
\end{abstract}

\section{Introduction}
\label{sec:intro}

Recently, convolutional neural networks (ConvNets) have made great contributions to image recognition. The current area of interest has been shifting toward temporal analysis of video input, such as action recognition and motion representation. These techniques are expected to be useful in such areas as visual surveillance, autonomous driving, and VR/AR. Various video recognition frameworks have been proposed to create a de facto standard for the field of motion representation. 

In this era of ConvNets, two-stream ConvNets~\cite{SimonyanNIPS2014} have replaced most of the hand-crafted approaches used on human action datasets. Two-stream ConvNets efficiently categorize human actions using the RGB image (spatial stream) and flow image (temporal stream), and they perform better than either dense trajectories (DTs) or improved dense trajectories (iDTs)~\cite{WangCVPR2011,WangICCV2013}.  We note that use of the spatial stream achieved over 70\% on the UCF101 dataset~\cite{UCF101}; that is, the per-frame RGB feature was able to categorize the 101 action classes without a temporal feature. The action datasets mainly occupy background areas behind the human areas in the image sequences.

\begin{figure}[t]
\begin{center}
    \includegraphics[width=0.7\linewidth]{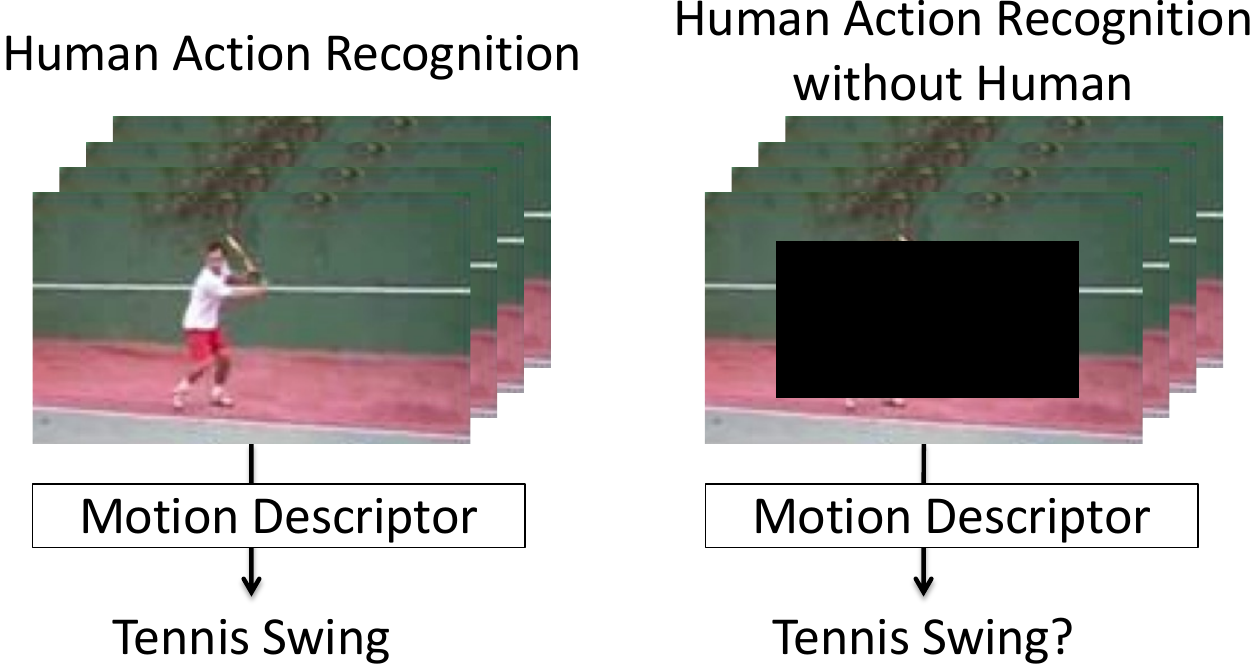}
\end{center}
   \caption{In our pre-experiment, we revealed that \textit{humanless} human action recognition can be done with a motion representation (e.g., two-stream ConvNets~\cite{SimonyanNIPS2014}). An undesirable scenario is arising because recent ConvNets rely heavily on a context-based representation. The problem of human action recognition with a video sequence is replaced by simple scene recognition.}
\label{fig:withouthuman}
\end{figure}

We believe that the problem of human action recognition in videos has been replaced by simple scene recognition without any temporal information. Because of high-level learning representations, ConvNets recognize action and context (e.g., object, scene) indiscriminately. This is why context can cheat human action recognition, as shown in Figure~\ref{fig:withouthuman}. We verified \textit{humanless} human action recognition with a spatial stream from an RGB sequence~\cite{SimonyanNIPS2014}. This simple trick is sufficient to understand a human action in video. We further analyze the context effects in videos to be clear that there is a lack of human action recognition. Moreover, it is important to find a more sophisticated way to represent motion.

We here identified two important issues:
\begin{itemize}
\item Evaluating context effects in video-based categorization.
\item Analyzing pure human motion for better motion representation.
\end{itemize}
We must arrange a couple of datasets to clarify the issues , including datasets , forcontext effects and pure human motion .


\begin{figure}[t]
  \centering
  \begin{subfigure}{0.45\linewidth}
    \centering
    \includegraphics[width=\linewidth]{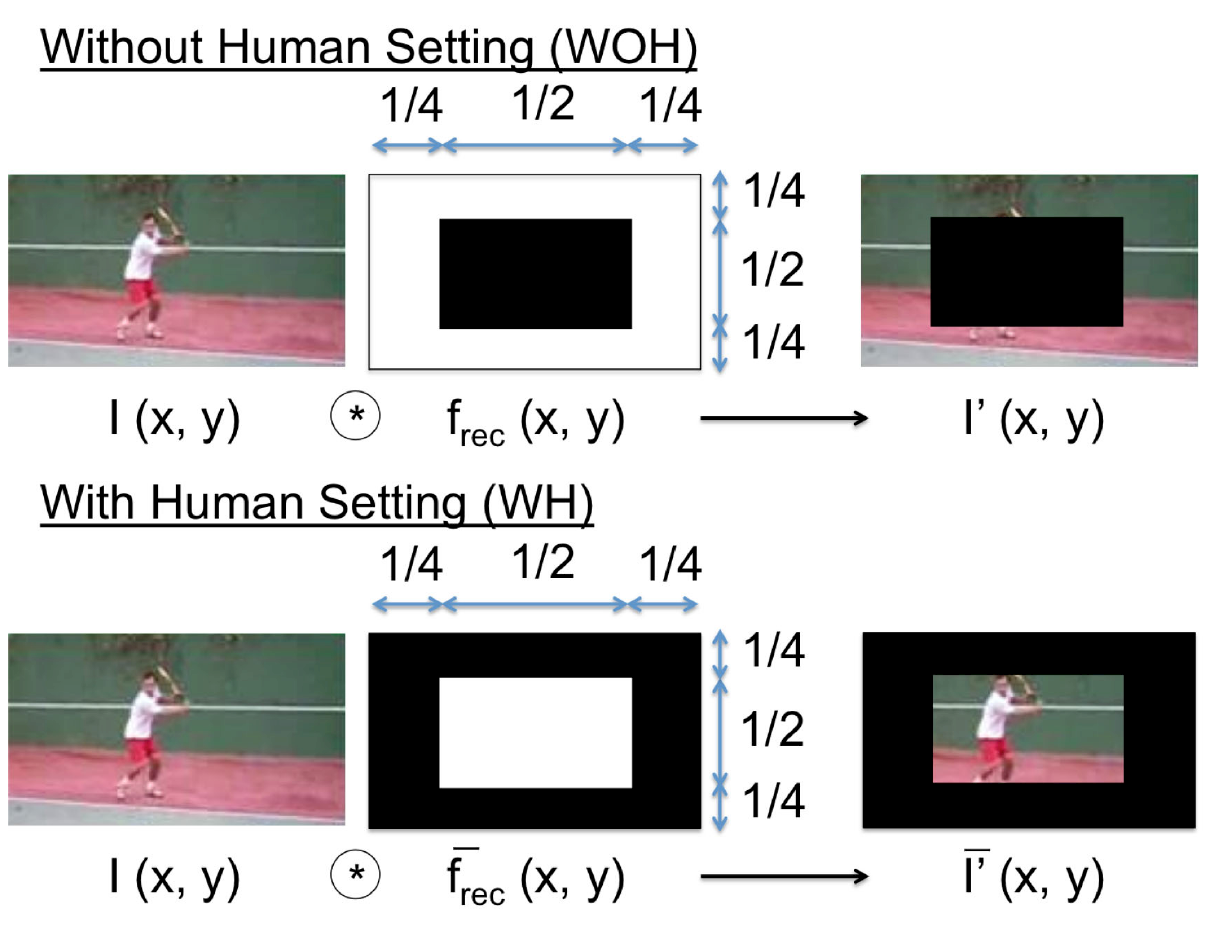}
    \caption{Rectangular image filters}
    \label{fig:filter}
  \end{subfigure}
  \hfill
  \begin{subfigure}{0.45\linewidth}
    \centering
    \includegraphics[width=\linewidth]{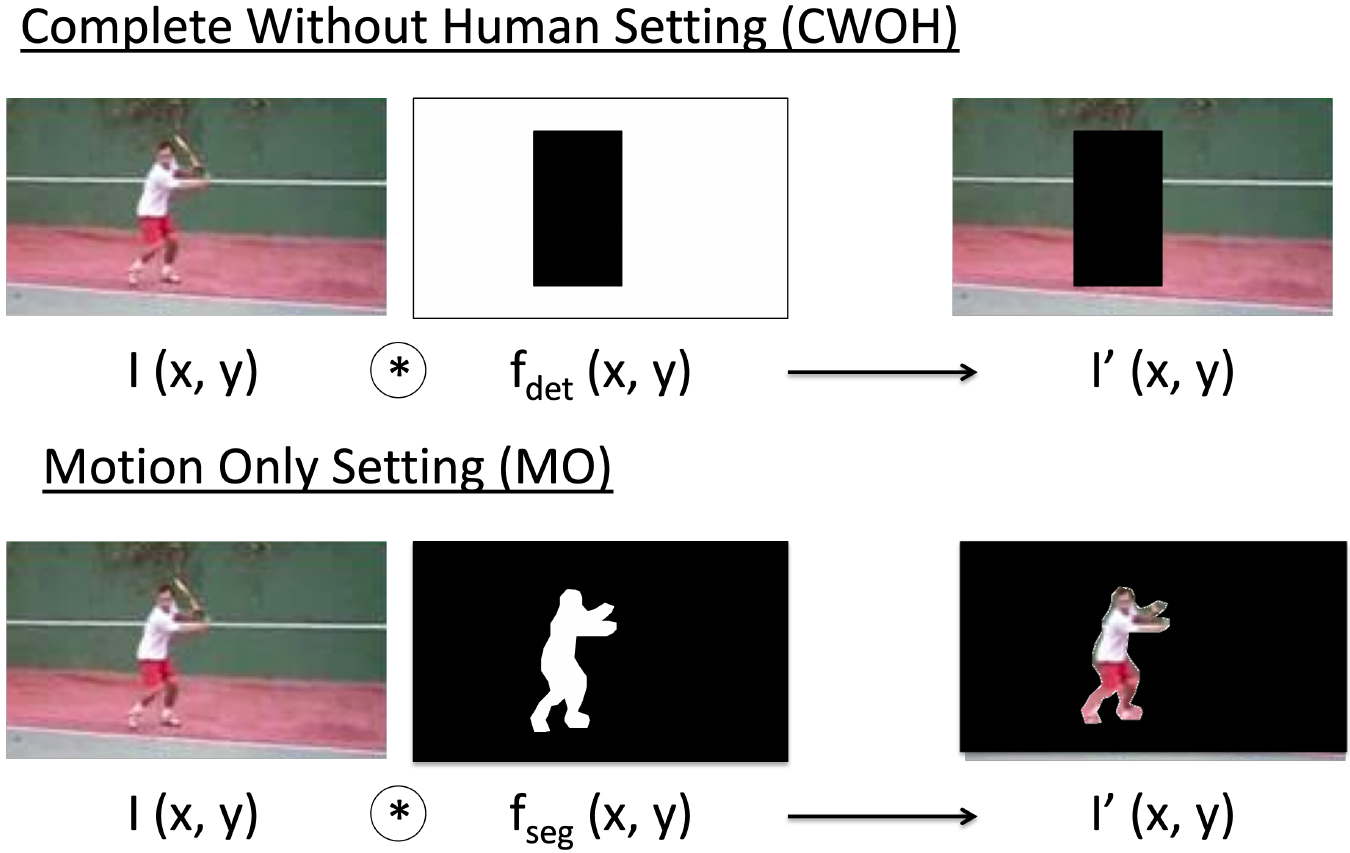}
    \caption{Improved image filters with human segmentation and object detection}
    \label{fig:improvedfilter}
  \end{subfigure}
  \caption{Four settings to evaluate background effects and pure human motions.}
  \label{fig:filters}
\end{figure}

In this paper, we evaluate background effects and pure human motions to show a lack of human action recognition in recent work . We build on the without-human setting by human detection and a motion-only setting with human segmentation (see Figure~\ref{fig:improvedfilter}) in addition to a simple rectangular-area rejection (see Figure~\ref{fig:filter}). Based on the two-stream ConvNets~\cite{SimonyanNIPS2014}, we separately analyze responses from both the spatial and temporal stream. In addition, to compare the performances of pure human motions with a couple of motion representations, we used existing representations for the motion-only datasets (MO-UCF101, MO-HMDB51), which were generated from \{UCF101, HMDB51\}.

We summarize our contributions as follows:
\begin{itemize}
\item We clearly show context effects on human action datasets. The motion representation in current use heavily relies on a feature that represents the background sequence.
\item Pure human motion is evaluated using the motion-only datasets, which consist of segmented humans.
\item We propose modified action datasets based on the UCF101~\cite{UCF101} and HMDB51~\cite{HMDB51} datasets. We created motion-only \{UCF101, HMDB51\} datasets (MO-UCF101, MO-HMDB51) in order to evaluate various approaches.
\end{itemize}

\section{Related work}

\subsection{Motion representation}

Space-time interest points (STIPs) were a first step in effectively treating space-time information in $xyt$ space~\cite{LaptevICCV2003,LaptevIJCV2005}. Improved versions of STIP~\cite{LaptevCVPR2008},~\cite{MarszalekCVPR2009},~\cite{EvertsCVPR2013} have been reported in studies in the field of action recognition. It has been claimed that space-time features are a necessary addition to STIPs, and this has produced, e.g., HOD3D and HOF. By evaluating st-points and descriptors~\cite{WangBMVC2009}, Wang \textit{et al.}~\cite{WangCVPR2011} turned dense sampling into a space-time feature. DTs are the trajectories that track densely sampled feature points. Additionally, Wang \textit{et al.}~\cite{WangICCV2013} proposed iDTs, in which the camera motion is estimated in order to remove the detection-based noise.

Several temporal models that use ConvNets have been proposed~\cite{TranICCV2015,SimonyanNIPS2014,WangCVPR2015,WangECCV2016,CarreiraCVPR2017}. Tran~\textit{et al.}~\cite{TranICCV2015} proposed 3D convolutional networks (C3D) with trained convolution filters on an $xyt$ space. 

Above this line , two-stream ConvNets is a well-organized algorithm that captures information from spatial and temporal streams~\cite{SimonyanNIPS2014}. The combination of RGB and flow information allows the performance of human action recognition to be significantly improved. The information obtained from the spatial stream is added to information from the temporal stream in order to classify action videos. Two approaches to this have been successful~\cite{WangCVPR2015,WangECCV2016}. These representations are based on using two-stream ConvNets to capture a motion feature using convolutional maps and network weights. Trajectory-pooled deep-convolutional descriptors (TDD) are at the intersection of iDT and two-stream ConvNets~\cite{WangCVPR2015}. Temporal segment networks (TSN) use a segmented image sequence to optimize the network parameters and evaluate the scene in each segment~\cite{WangECCV2016}. Using sophisticated video annotation and a large number of GPUs, two-stream 3D ConvNets improved performance on frequently used datasets, such as UCF101 (98.0\%) and HMDB51 (80.7\%)~\cite{CarreiraCVPR2017}.

The current ConvNet-based approaches rely heavily on a two-stream architecture consisting of a spatial stream and a temporal stream. We suspect that the spatial stream obtained by using RGB data will contain much contextual information, such as that from background scenes and objects. By using two-stream ConvNets, the spatial stream can be used to resolve the problem of action recognition (72.8\% with the original two-stream ConvNets~\cite{SimonyanNIPS2014}, 82.8\% with TDD~\cite{WangCVPR2015}, 84.5\% with TSN~\cite{WangECCV2016}, and ultimately, 95.6\% with two-stream 3D ConvNets~\cite{CarreiraCVPR2017}).

\subsection{Context model for understanding an action}

Contextual approaches focus on the region around a human and provide important cues that can be used for sophisticated human action recognition. In recent work, Jain \textit{et al.}~\cite{JainCVPR2015}, Wu \textit{et al.}~\cite{WuCVPR2016}, and Wang~\textit{et al.}~\cite{LiminWangIJCV2017} showed that object and scene context assist in the recognition of human actions. 

Jain \textit{et al.}~\cite{JainCVPR2015} evaluated how much of an object needed to be used in order to recognize the action. They combined object information with a classifier score to represent the area of a human. A large number of object labels, such as computer and violin, were used as priors and were assigned an output function.
%
%
An impressive study was that of Wu \textit{et al.}~\cite{WuCVPR2016}, in which responses from combined ConvNet architectures were employed by learning on an object and scene dataset. They used an augmented ImageNet~\cite{RussakovskyIJCV2015} and Places205~\cite{Places205} dataset for training a fusion network. When the optimized network is used in conjunction with certain features , it performs much better than other video understanding methods, including iDT and plain ConvNet. Wang \textit{et al.}~\cite{LiminWangIJCV2017} presented a similar study with ImageNet and Places, but they transferred from pre-trained object and scene parameters to high-level event features. They effectively utilized general features to understand events including human actions with a still image.
 
According to these results, the responses enhanced by objects and scenes seem to be very strong when used with the current action recognition datasets. To confirm the background effects and pure human motions, we tried to verify the \textit{human action recognition without a human} approach with a sophisticated setting .

\section{Evaluation of background effects and pure human motions}
\label{sec:withouthuman}

\textbf{Summary.} We carry out experiments using the human action recognition without a human approach, based on the four settings shown in Figure~\ref{fig:filters}: (i) without-human setting (WOH), (ii) with-human setting (WH), (iii) complete without-human setting (CWOH), and (iv) motion-only setting (MO). We would like to clarify the effects of the background and pure human motions. The very deep two-stream ConvNets~\cite{WangarXiv2015} was used as the baseline algorithm. We employ the UCF101~\cite{UCF101} and HMDB51~\cite{HMDB51} datasets, which are the ones most frequently used for human action recognition.

\textbf{Without-human setting (WOH; see Figure~\ref{fig:filter} top).} In the WOH setting, we calculated the image filtering with a black background, as follows:

\begin{eqnarray}
   I^{'}(x,y) &=& I(x,y) * f_{rec}(x,y),
\end{eqnarray}
where $I^{'}$ and $I$ are the filtered and input images, respectively,  and $x,y$ are pixel elements. Filter $f_{rec}$ replaces the center-around area with a black rectangle. The detailed operation is shown at the top of Figure~\ref{fig:filter}.

\textbf{With-human setting (WH; see Figure~\ref{fig:filter} bottom).} The image from the WH setting is shown as follows: 

\begin{eqnarray}
   \overline{I^{'}}(x,y) &=& I(x,y) * \overline{f_{rec}}(x,y),
\end{eqnarray}
where $\overline{I^{'}}$ is the inverse image of $I^{'}$, and $\overline{f_{rec}}$ is the filter $f_{rec}$ in the WOH. The background area is eliminated with the inverse filter, as shown at the bottom of Figure~\ref{fig:filter}.

The drawback of both WOH and WH is that the simple filter cannot hide/display a human area. We applied human detection and segmentation to UCF101 and HMDB51.

\textbf{Complete without-human setting (CWOH; see Figure~\ref{fig:improvedfilter} top)} We employed a human detection approach in order to reject textures around a human. Compared to the WOH setting, the CWOH can evaluate a wider background area in an image sequence. The CWOH should not be replaced by human segmentation, since a shadow in the segmented human area can be a motion cue.

The CWOH is as follows:

\begin{eqnarray}
   I_{det}^{'} &=& I(x,y) * f_{det}(x,y),
\end{eqnarray}
where $f_{det}$ includes the detected bounding box $b$ as

\begin{eqnarray}
   f_{det}(x,y) &=& \Sigma_{i}^{N}
   \left\{
   \begin{array}{l}
   0, \;\; (if\; (x,y)\; is\; in\; the\; bbox.) \\
   1, \;\; (otherwise.)\\
   \end{array}
   \right.
\end{eqnarray}
in which $N$ is the number of bounding boxes per frame.

The CWOH is effective in helping us understand the effects of the background. The details of the CWOH dataset are described in Section~\ref{sec:cwohdb}.

\begin{table}[t]
\begin{center}
\begin{tabular}{lcc}
\hline
Stream & UCF101 & HMDB51 \\
\hline\hline
Two-stream & 84.30 & 59.02 \\
Spatial stream & 74.86 & 36.60 \\
Temporal stream & 80.33 & 56.73 \\
\hline
\end{tabular}
\end{center}
\caption{Baseline performance rates on the UCF101 and HMDB51 datasets with two-stream ConvNets~\cite{WangarXiv2015}.}
\label{tab:baseline}
\end{table}

\textbf{Motion-only setting (MO; see Figure~\ref{fig:improvedfilter} bottom)} To represent a pure motion in a human area, we use a mask for human segmentation; in this case, we use a semantic segmentation. Unlike the WH setting, some background areas are contained in an image sequence, and so we generate a human-centric dataset for pure human motion analysis. The details of MO datasets are described in Section~\ref{sec:modb}.

The image of the MO is as follows:

\begin{eqnarray}
   I_{seg}^{'} &=& I(x,y) * f_{seg}(x,y),
\end{eqnarray}
where filter $f_{seg}$ is the result of pixel-wise human segmentation. The background is eliminated with a segmented filter at the bottom of Figure~\ref{fig:improvedfilter}.

\textbf{Baseline and training with very deep two-stream ConvNets.} We used very deep two-stream ConvNets~\cite{WangarXiv2015} as a baseline. A 16-layer VGGNet was assigned for each stream in the architecture. The spatial stream outputs a probabilistic distribution from a 224 pixel $\times$ 224 pixel $\times$ 3 channel image, and the temporal stream was obtained from a stacked flow input that was 224 pixels $\times$ 224 pixels $\times$ 20 channels. The model is trained with UCF101, and additional training was performed with HMDB51. By using the pretrained models for UCF101 and HMDB51, we obtained the results shown in Table~\ref{tab:baseline}. Our implementation is different from that reported by Wang~\cite{WangarXiv2015} on UCF101. The performance rate (mean average precision; mAP) depends on the parameter tuning. They reported 78.4\% (spatial), 87.0\% (temporal), and 91.4 \% (two-stream) with UCF101. The original study did not report a rate for HMDB51; however, the results of our tuning (59.02\%) were comparable to those of the state-of-the-art models (around 60\% for that dataset).

We then updated the network parameters with HMDB51 by fine-tuning with ConvNet. The learning procedure of the spatial and the temporal streams is based on that of a previous study~\cite{WangarXiv2015}. The initial learning  rate was set to 0.001, with an updating rate of 0.1 for every 10,000 iterations. The learning of the acceleration stream terminates after 10,000 (spatial) and 30,000 (temporal) iterations. We assigned a high dropout ratio in each of the fully connected (fc) layers: 0.8 for the first and 0.9 for the second.

\section{Augmented datasets}
\label{sec:motiondb}
\begin{figure}[t]
\begin{center}
   \includegraphics[width=0.75\linewidth]{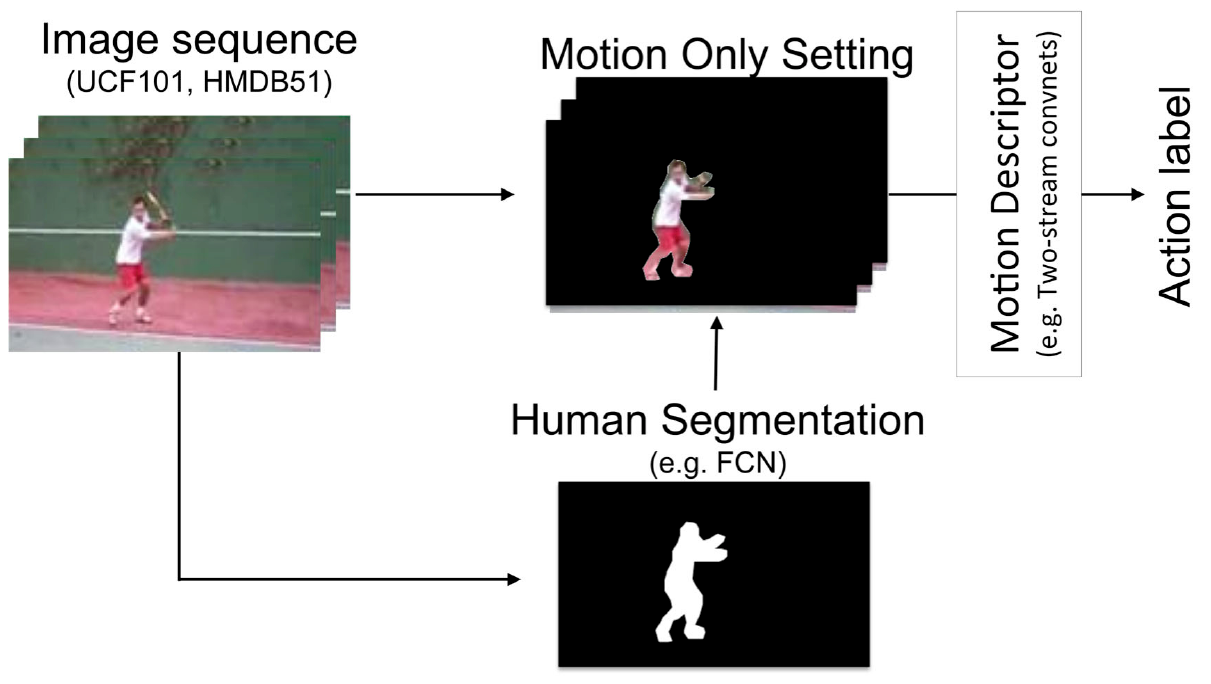}
\end{center}
   \caption{Flowchart of dataset creation: We created motion-only \{UCF101,HMDB51\} datasets for evaluating pure human motions. They were also used to confirm that the motion representation used in existing approaches is reliable. A trial was performed to clarify the strategy for developing a sophisticated motion representation.}
\label{fig:howtodb}
\end{figure}

\begin{figure}[t]
  \centering
  \begin{subfigure}{1.0\linewidth}
    \centering
    \includegraphics[width=\linewidth]{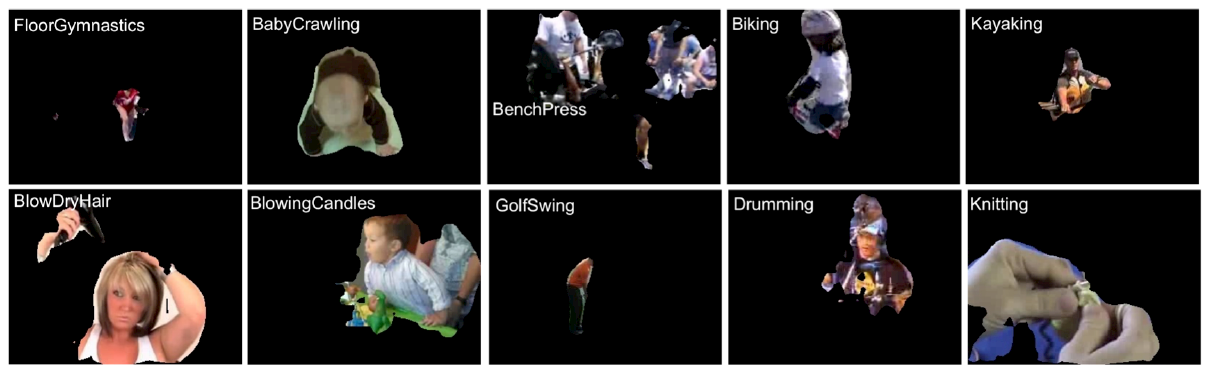}
    \caption{Motion-only UCF101 dataset}
    \label{fig:moucf101}
  \end{subfigure}
  \\
  \begin{subfigure}{1.0\linewidth}
    \centering
    \includegraphics[width=\linewidth]{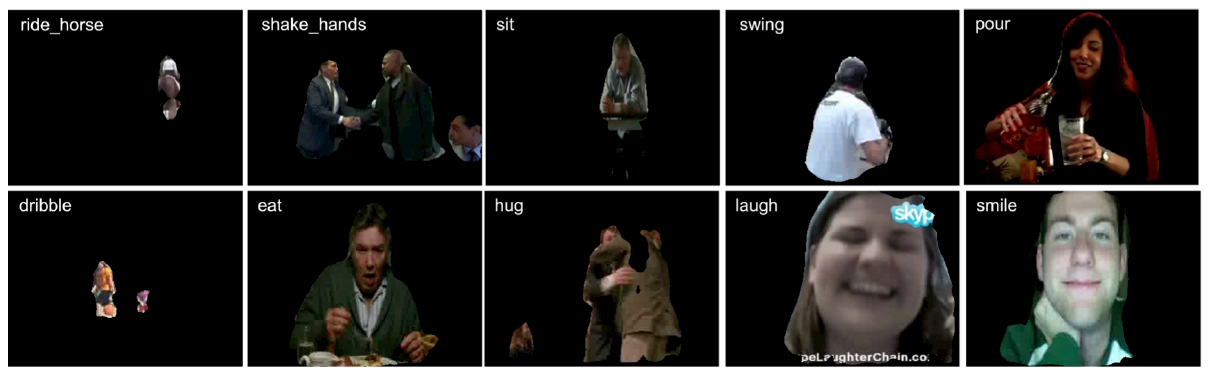}
    \caption{Motion-only HMDB51 dataset}
    \label{fig:mohmdb51}
  \end{subfigure}
  \caption{Motion-only \{UCF101, HMDB51\} datasets with a semantic segmentation. Each dataset must be created from a good trial to represent pure motion.}
  \label{fig:motiononlydb}
\end{figure}

We have generated augmented datasets based on UCF101~\cite{UCF101} and HMDB51~\cite{HMDB51}. Figure~\ref{fig:howtodb} shows the flowchart for creating the datasets, and Figure~\ref{fig:motiononlydb} shows examples of the use of motion-only UCF101 and HMDB51 to extract a pure motion from the eliminated backgrounds. Detailed descriptions of these components follow.

\subsection{Original datasets.} 

First, we describe the UCF101~\cite{UCF101} and HMDB51~\cite{HMDB51} datasets, which have been extensively studied in action recognition applications. 

UCF101 was mainly collected from YouTube videos of sports and musical instrument performance scenes. The recognition task is to predict an action label from a given video. The dataset contains several difficulties with computer vision, such as camera motion, scaling, posture changes, and viewpoint differences. The data provided three divided train/test splits, which were used to calculate a cross-validated mean average precision (mAP). The data contained a total of 13,320 videos, of which 10,000 were used for training and 3,000 were used for testing. 

HMDB51 is a collection of movies from various Internet video services, such as the Prelinger archive, YouTube, and Google Videos. The objective is to review a large number of video clips, and so the dataset contains 6,849 short clips divided into 51 action classes. Each action class contains a minimum of 101 video clips. The 51 action classes are categorized into 5 groups.

\begin{figure}[t]
  \centering
  \begin{subfigure}{0.24\linewidth}
    \centering
    \includegraphics[width=\linewidth]{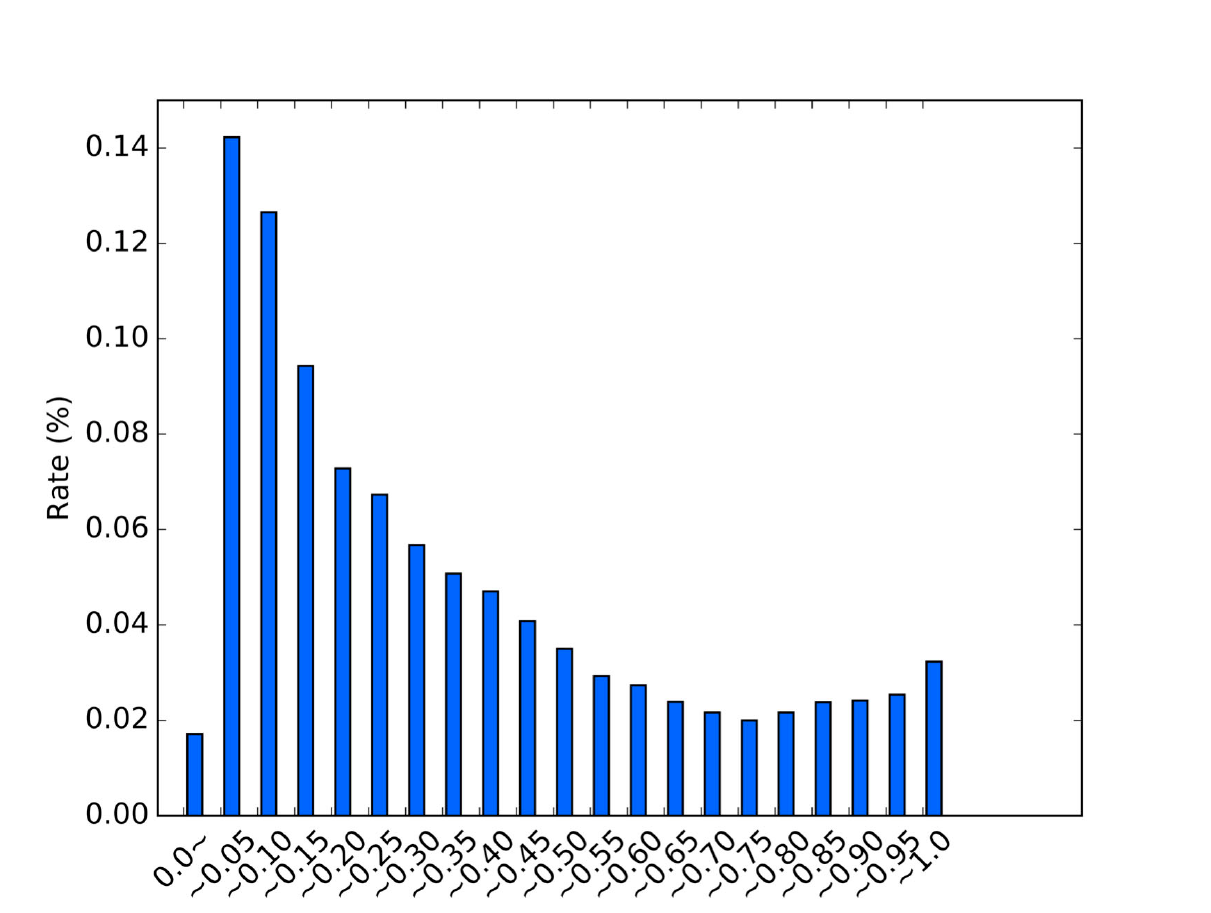}
    \caption{Percent of total area that is in the rectangle; CWOH-UCF101 (average: 0.323)}
    \label{fig:cwoh_ucf101}
  \end{subfigure}
  \hfill
    \begin{subfigure}{0.24\linewidth}
    \centering
    \includegraphics[width=\linewidth]{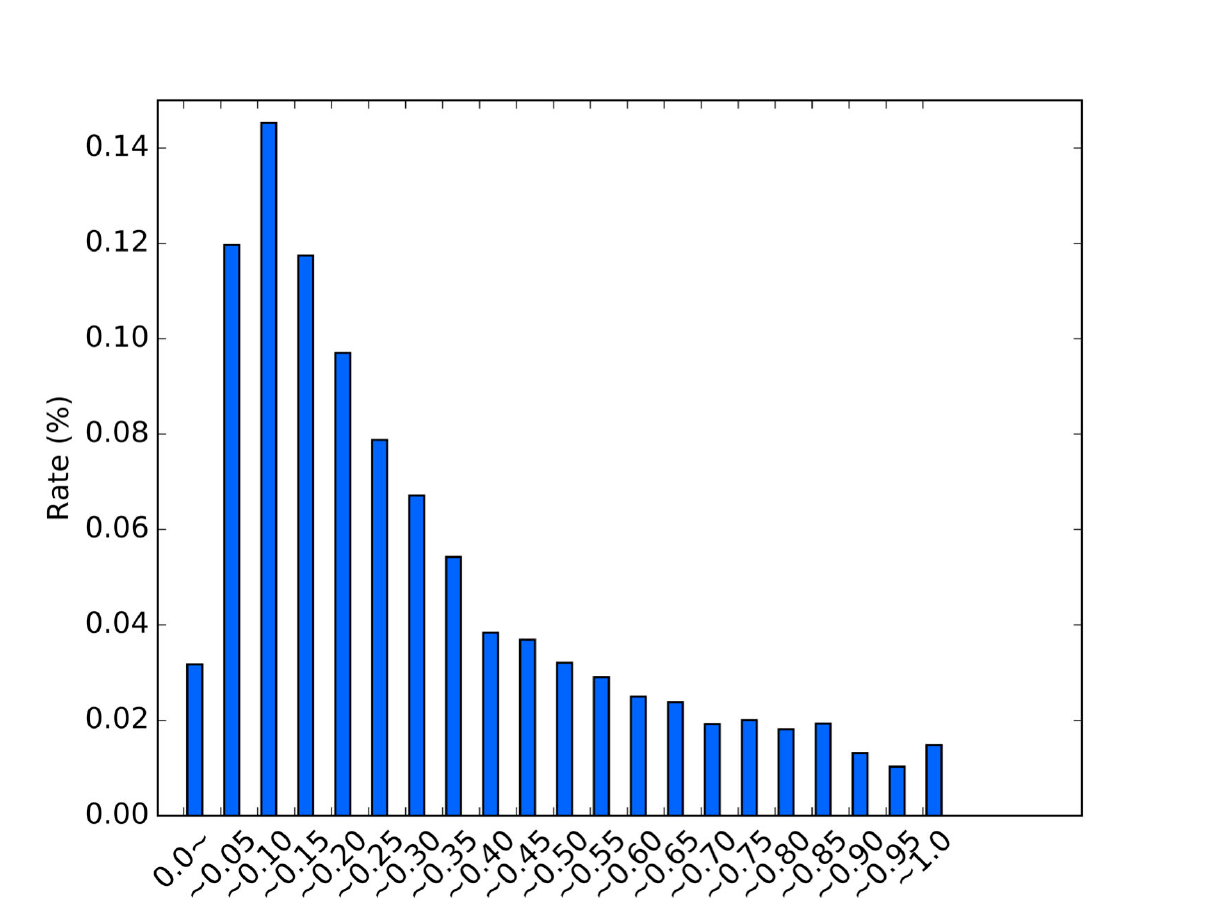}
    \caption{Percent of total area that is in the rectangle; CWOH-HMDB51 (average: 0.278)}
    \label{fig:cwoh_hmdb51}
  \end{subfigure}
  \hfill
  \begin{subfigure}{0.24\linewidth}
    \centering
    \includegraphics[width=\linewidth]{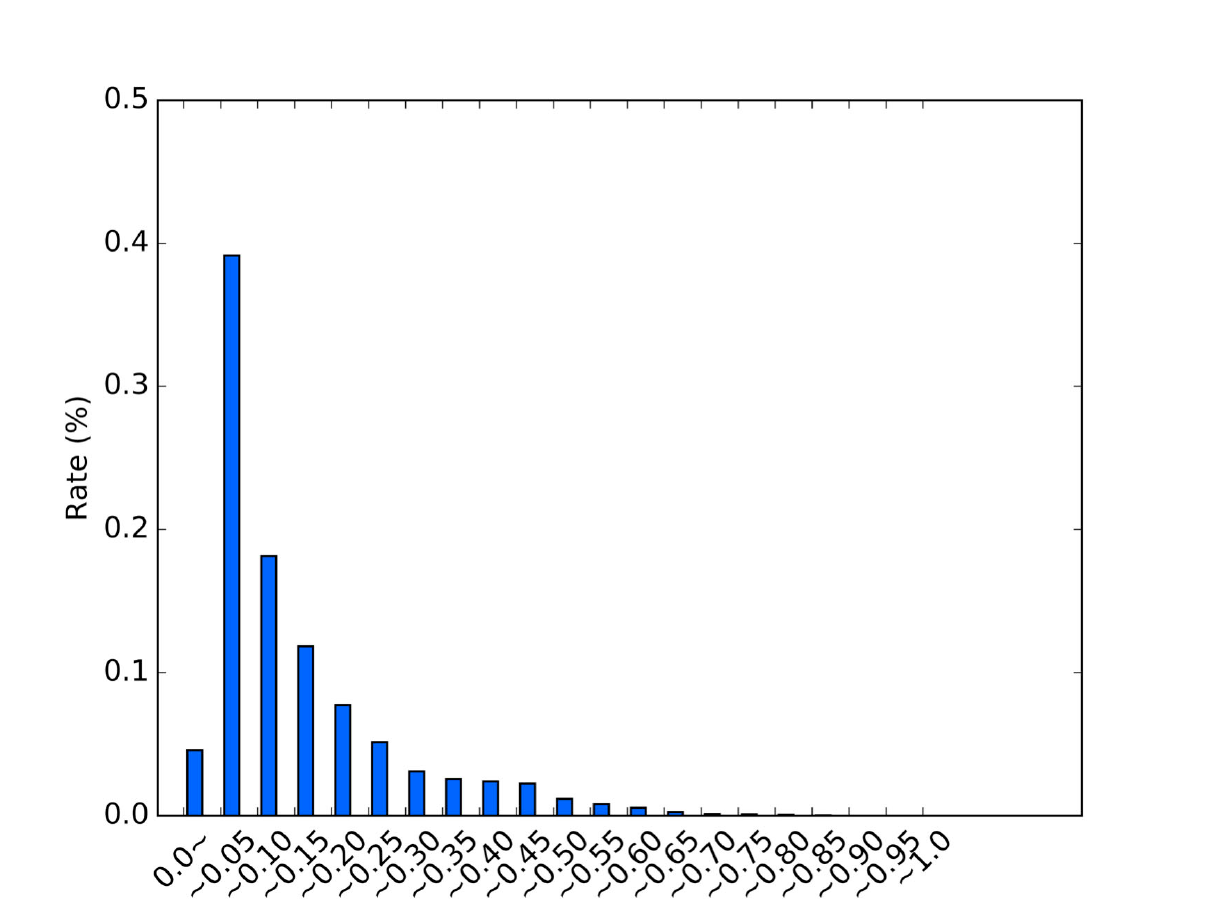}
    \caption{Statistics of MO-UCF101: \# human pixels / all (average: 0.112)}
    \label{fig:mo_ucf101}
  \end{subfigure}
  \hfill
  \begin{subfigure}{0.24\linewidth}
    \centering
    \includegraphics[width=\linewidth]{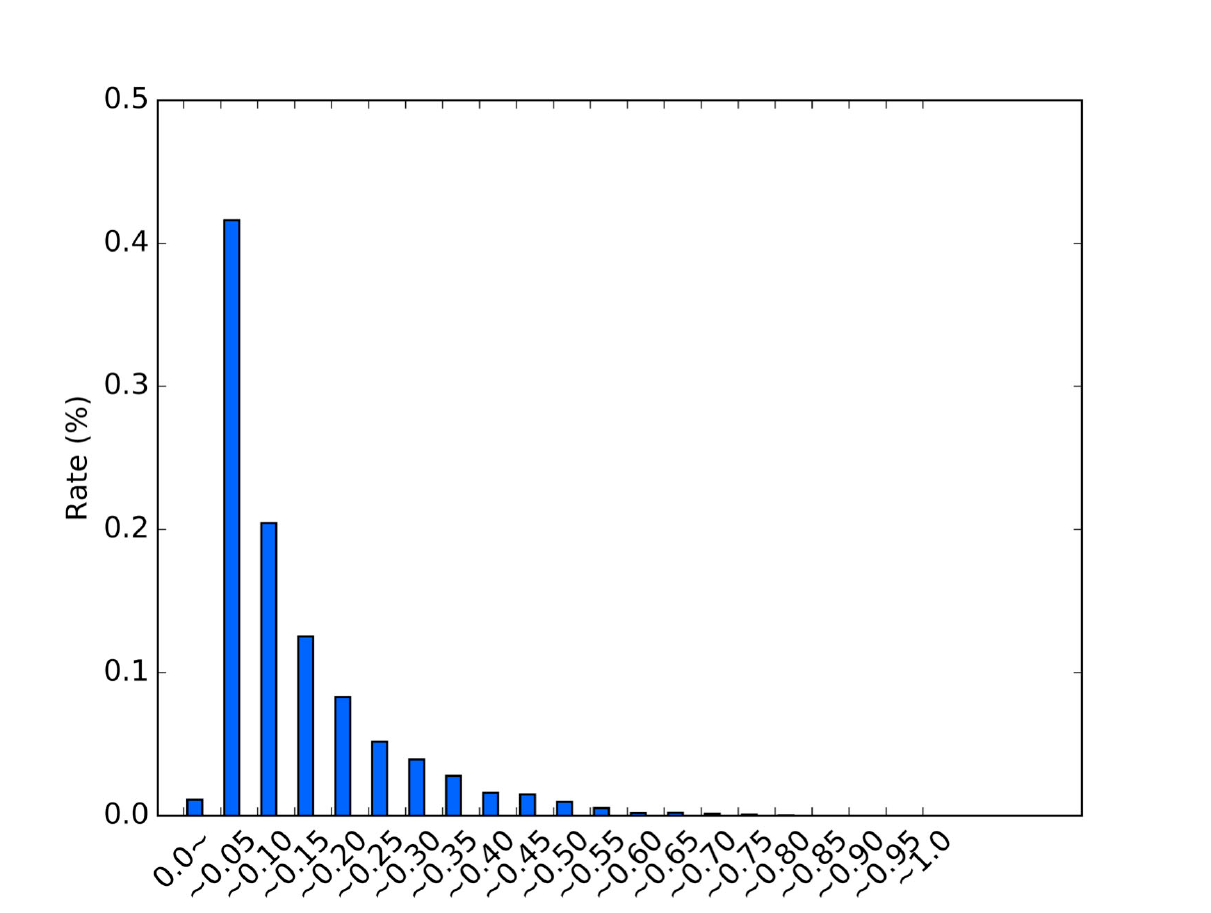}
    \caption{Statistics of MO-HMDB51: \# human pixels / all (average: 0.107)}
    \label{fig:mo_hmdb51}
  \end{subfigure}
  \caption{Area statistics of the CWOH and MO datasets: For the CWOH setting, the value given is the percentage of all pixels that are in the rectangular area, and for the MO setting, it is the percentage of all pixels that are in the segmented human area.}
  \label{fig:statsdb}
\end{figure}

\subsection{Complete without-human \{UCF101, HMDB51\} datasets (CWOH-UCF101, CWOH-HMDB51)}
\label{sec:cwohdb}
The datasets were developed to investigate the effects of the background on action recognition. We used the faster R-CNN~\cite{RenNIPS2015} to detect humans and reject the textures around them. 
The thresholding score was fixed at 0.8. In addition, we manually replaced the detections when the detection was incorrect ~\footnote{Best effort.}.

\subsection{Motion-only \{UCF101, HMDB51\} (MO-UCF101, MO-HMDB51; Figure~\ref{fig:howtodb}).}
\label{sec:modb}
We created augmented datasets in order to better evaluate pure human motions. The basic training and testing procedures were the same as those used on the original datasets. To create a human-centered area, we employed fully convolutional networks (FCN)~\cite{LongCVPR2015}, which are the premier method for semantic segmentation. We used the Pascal VOC pretrained model provided by the authors. 
The ``person" category in the FCN was retained in the motion-only datasets. In both datasets, the human-centered semantic segmentation worked well when evaluating pure motions (see Figure~\ref{fig:motiononlydb}).

This was a significant result for pure human motion representation. We updated the parameters in the baseline two-stream ConvNets with the motion-only datasets. Moreover, we compared the results with existing motion representations, such as iDT, two-stream ConvNets, and C3D, using the MO-UCF101 and MO-HMDB51 datasets. Details of these experiments are shown in Section 5.

\subsection{Dataset statistics}

Figure~\ref{fig:statsdb} shows statistics for CWOH-UCF101, CWOH-HMDB51, MO-UCF101, and MO-HMDB51. Figure~\ref{fig:cwoh_ucf101} and~\ref{fig:cwoh_hmdb51} show the percentage of all pixels that are in the rectangular area. Figure~\ref{fig:mo_ucf101} and~\ref{fig:mo_hmdb51} show the percentage of all pixels that are in the segmented area. Although these values are similar, the HMDB51 videos tend to have smaller areas that are interpreted as humans . The average proportion of the pixels that are parts of humans are as follows: 0.323 for CWOH-UCF101, 0.112 for MO-UCF101, 0.278 for CWOH-HMDB51, and 0.107 for MO-HMDB51. The property makes HMDB51 difficult in the performance rate .

\begin{table*}[t]
\begin{center}
\begin{tabular}{lcccc}
\hline
Dataset & WOH-UCF101 & WH-UCF101 & WOH-HMDB51 & WH-HMDB51 \\
\hline\hline
Two-stream & \textbf{47.42} & \textbf{56.91} & 17.06 & 14.97 \\
Spatial stream & 45.33 & 51.26 & \textbf{17.58} & 7.06 \\
Temporal stream & 26.80 & 40.50 & 9.41 & \textbf{16.01} \\
\hline
\end{tabular}
\end{center}
\vspace{-5.0pt}\caption{WOH and WH performance rates (\%) on the UCF101 and HMDB51 datasets.}
\label{tab:wh_woh}
\end{table*}

\begin{table*}[h!]
\begin{center}
\begin{tabular}{lcccc}
\hline
Dataset & CWOH-UCF101 & MO-UCF101 & CWOH-HMDB51 & MO-HMDB51 \\
\hline\hline
Two-stream & \textbf{53.07} & \textbf{59.37} & 5.16 & \textbf{28.17} \\
Spatial stream & 49.96 & 45.76 & 2.68 & N/A \\
Temporal stream & 39.30 & 56.36 & \textbf{7.58} & 28.17 \\
\hline
\end{tabular}
\end{center}
\vspace{-5.0pt}\caption{Performance rates (\%) on the CWOH-UCF101, MO-UCF101, CWOH-HMDB51, and MO-HMDB51 datasets.}
\label{tab:cwoh_mo}
\end{table*}

\begin{table*}[t]
\begin{center}
\begin{tabular}{lcc}
\hline
 & MO-UCF101 & MO-HMDB51 \\
\hline\hline
ResNet-C3D~\cite{TranarXiv2017} & \textbf{84.21} & N/A \\
Two-stream ConvNets~\cite{WangarXiv2015} & 59.37 & \textbf{28.17} \\
\hspace{10.0pt} Spatial stream & 45.76 & N/A \\
\hspace{10.0pt} Temporal stream & 56.36 & 28.17 \\
iDT (Combine)~\cite{WangICCV2013}& 54.23 & 28.06 \\
\hspace{10.0pt} iDT (HOG) & 31.86 & 22.03 \\
\hspace{10.0pt} iDT (HOF) & 26.09 & 15.40 \\
\hspace{10.0pt} iDT (MBH) & 29.47 & 20.05 \\
\hline
\end{tabular}
\end{center}
\vspace{-5.0pt}\caption{Performance rates (\%) for representations of motion with the motion-only \{UCF101, HMDB51\} dataset.}
\label{tab:comparison}
\end{table*}

\section{Experiments}

We carried out trials using the four scenarios discussed in Section~\ref{sec:withouthuman}: WOH, WH, CWOH, and MO. This section is separated into two subsections. The first presents experiments on the four settings. In the latter subsection, we will compare some recently developed approaches, including iDT~\cite{WangICCV2013} and ResNet-C3D~\cite{TranICCV2015,TranarXiv2017} on the MO-UCF101/MO-HMDB51 dataset, in addition to two-stream ConvNets~\cite{WangarXiv2015} as a baseline.

\subsection{Four settings: WOH, WH, CWOH, and MO}

We first summarize the  scenarios as follows:
\begin{itemize}
   \item With a fixed rectangle in WOH-UCF101 (see Table~\ref{tab:wh_woh}), two-stream ConvNets gave a performance of 47.42 and the spatial stream classified 45.33 in 101 action categories even though most of the human area was occluded. Moreover, more sophisticated CWOH setting increased the accuracies of both the two-stream ConvNets (53.07) and the spatial stream (49.96) in Table~\ref{tab:cwoh_mo}. We proved that the contextual features are sufficient to understand human action recognition.
   \item In contrast to the without-human settings, two-stream ConvNets decreased the performance rate from a baseline of 84.30@UCF101 to 56.91@WH-UCF101. The loss in the performance rate is due to a lack of context in an image sequence. Recent motion representation techniques indiscriminately recognize human actions together with their context , so a lack of context decreases an accuracy of human action recognition.
   \item It is more difficult to understand human actions with HMDB51 . Although human actions are categorized into 51 classes, the dataset has various backgrounds per action category since the daily actions occur frequently.
\end{itemize}
Table~\ref{tab:wh_woh} shows the performance rate on the UCF101 dataset for the WH and WOH approaches. The two-stream CNN had a rate of 47.42 with the WOH. With the WOH, the results for the spatial stream were +18.53 ($45.33 - 26.80$) better than those for the temporal stream. Therefore, the spatial stream derived from RGB input is able to classify human actions with backgrounds. The temporal stream contributes a small amount to the background classification; that is, the performance rate with the WOH is only increased by +2.09 ($47.42 - 45.33$) when the temporal stream is added. The two-stream CNN rate was 56.91 with the WH, which is only +9.49 ($56.91-47.42$) higher than that with the WOH. 

We also evaluated the effect of the WOH and WH on HMDB51. We were surprised to find that the WOH (17.06) result was higher than that of the WH (14.97); that is, the background feature was more effective than the human-centered feature. It was also clearly shown that with the WOH, the results with the spatial stream (17.58) were better than those with the temporal stream (9.41), and with the WH, the results with the temporal stream (16.01) were better than those with the spatial stream (7.06).

These results show that the spatial stream in the two-stream CNN uses a background sequence to classify human actions in both the UCF101 and HMDB51 datasets. The reason for this is that the spatial stream relies heavily on the weights that were originally trained on the ImageNet. The ImageNet training allows it to easily understand background objects and scenes.

Table~\ref{tab:cwoh_mo} shows the more detailed human action analysis obtained with the CWOH and MO. The results of the CWOH-UCF101 were +5.65 ($53.07 - 47.42$) better than those of the WOH-UCF101, and those of the CWOH-HMDB51 were $-11.90$ ($5.16 - 17.06$) worse than those of the WOH-HMDB51. The CWOH dataset contains a greater amount of background in each image sequence. The performance rate of CWOH depends on the background effects. The results indicate that the CWOH-UCF101 has a greater reliance on the background features when performing a classification. The CWOH-UCF101 backgrounds are more easily classified, since they primarily consist of sports videos. Compared to the CWOH-UCF101, it is more difficult to classify the background features of the CWOH-HMDB51, because the dataset includes a wider variety of videos, including scenes such as daily living and human-human interactions. The spatial stream of each dataset is an indicator of the degree of difficulty of interpreting the background features: 49.96 with CWOH-UCF101 and 2.68 with WOH-HMDB51. 

Next, we consider MO-UCF101 and MO-HMDB51 (see Table~\ref{tab:cwoh_mo}). The most important problem is that of representing a pure human motion in an image sequence. This trial allowed us to consider only action recognition, independent of interpreting the background. The two-stream CNN had a performance of 59.37 with MO-UCF101 and 28.17 with MO-HMDB51; in both cases, the temporal stream results were better than those of the spatial stream. The MO setting results in an improved performance on each of the datasets, compared to the results of the WH setting: +2.46 ($59.37 - 56.91$) with UCF101 and +13.20 ($28.17 - 14.97$) with HMDB51. Unlike the WH, which contains a human motion and a partial background, the MO contains a pure human motion. The results show that it is easier to optimize the temporal stream for the pure human motions. In the case of a spatial stream, we easily optimized MO-UCF101 (45.76); however, MO-HMDB51 was not converged since unstable areas frequently appeared between humans and background. Subsection~\ref{sec:puremotion} presents a detailed comparison of the existing approaches to representing motion.

\subsection{Pure motion representation}
\label{sec:puremotion}
We investigated the effectiveness of various motion representations, including iDT~\cite{WangICCV2013} and ResNet-C3D~\cite{TranICCV2015,TranarXiv2017}; we included the two-stream ConvNets~\cite{WangarXiv2015} as a baseline.

\textbf{iDT~\cite{WangICCV2013}.} iDT is the de facto standard hand-crafted model for video representation. The setting is based on the original implementation. To generate a codeword vector, MBH (192-d), HOF (108-d), and HOG (96-d) are captured each time a trajectory is sampled; they are incorporated into a feature vector.

\textbf{ResNet-C3D~\cite{TranICCV2015,TranarXiv2017}.} C3D employs a 3D convolutional filter on an $xyt$ space and is generated from an RGB sequence~\cite{TranICCV2015}. Moreover, the improved C3D with ResNet-18 architecture is used on the MO datasets. Fine-tuning is implemented for each MO dataset in addition to the Sports-1M pre-trained ResNet-C3D.

Table~\ref{tab:comparison} shows the performance rates for various motion representations with the MO-UCF101 and MO-HMDB51 datasets. The ResNet-C3D achieved the best rate of 84.21 on the MO-UCF101 dataset. Although Res3D was not converged on MO-HMDB51, a sharp feature extraction with skip-connection and spatiotemporal 3D convolution is beneficial for motion-based UCF101. There are some differences between MO-UCF101 and MO-HMDB51; for example, MO-UCF101 and MO-HMDB51 contain sports and daily videos, respectively, and MO-UCF101 is two times bigger than MO-HMDB51 in terms of \#video. MO-UCF101 tends to clearly segment human areas because the dataset contains a lot of sports videos, which tend to be limited to human-centered videos . Moreover, the amount of video data in MO-UCF101 is easy to converge in the residual 3D networks. The data properties require us to train a sophisticated motion representation for MO-UCF101 with pure human motions. Further, it is relatively easier to optimize a ConvNet model with MO-UCF101. The second-best rate was two-stream ConvNets (59.37@MO-UCF101, 28.17@MO-HMDB51) but the temporal stream contributed greatly to the performance rate, with 56.36@MO-UCF101 and 28.17@MO-HMDB51. These values proved that motion features like optical flow are also useful on the motion-only datasets.

According to the experimental result on MO-UCF101 in Table~\ref{tab:comparison}, a combination of 3D ConvNet and ResNet is the best (84.21), and the second best is optical flow representation with a temporal stream (56.36) and iDT (54.23). 
ResNet-C3D enables a significant representation to be caught for MO-UCF101. This is a useful area for future research.

\begin{figure}[t]
  \centering
  \begin{subfigure}{0.8\linewidth}
    \centering
    \includegraphics[width=\linewidth]{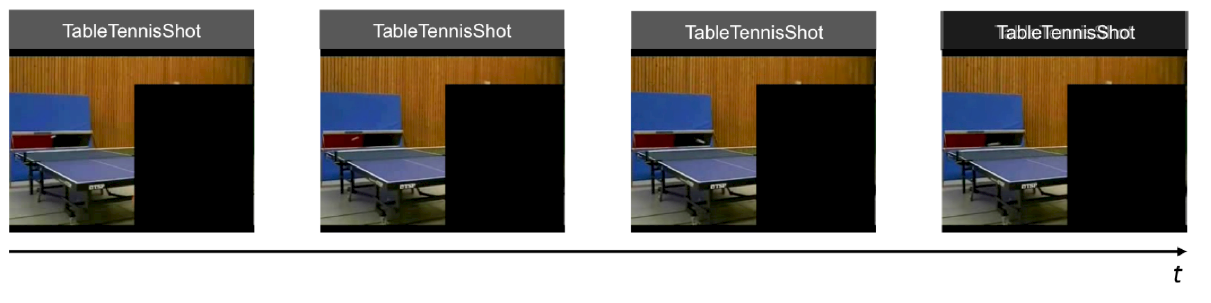}
    \caption{Complete without-human setting in \textit{table tennis shot}: The system outputs a correct answer.}
    \label{fig:woh001}
  \end{subfigure}
  \\
    \begin{subfigure}{0.8\linewidth}
    \centering
    \includegraphics[width=\linewidth]{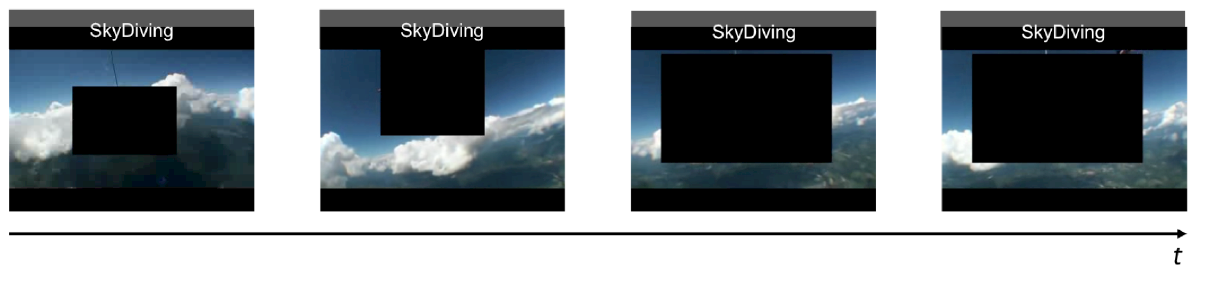}
    \caption{Complete without-human setting in \textit{skydiving}: The system outputs a correct answer.}
    \label{fig:woh002}
  \end{subfigure}
  \\
  \begin{subfigure}{0.8\linewidth}
    \centering
    \includegraphics[width=\linewidth]{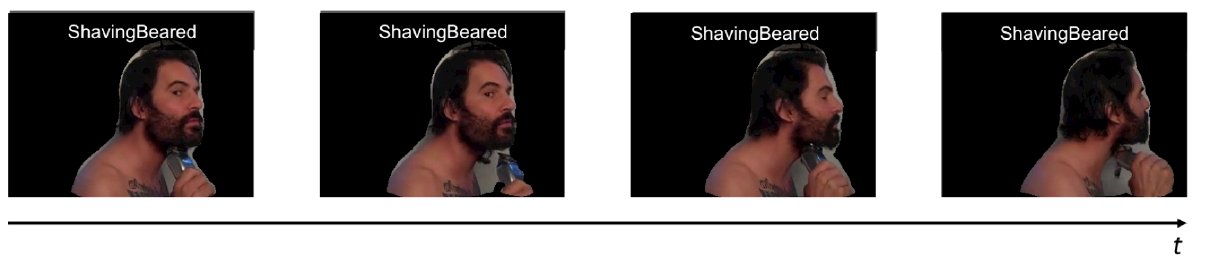}
    \caption{Motion only setting in shaving beard : The system outputs a correct answer.}
    \label{fig:woh003}
  \end{subfigure}
  \caption{Examples of complete without-human and motion-only settings. We showed that recent motion representation techniques can understand human action without a human (which is related to context ) and pure human motion. This allows us to improve the level understanding of human action.}
  \label{fig:woh_vis}
\end{figure}

We further implemented ResNet-C3D on CWOH-UCF101 with human detection. A surprising result of 65.48 is recorded for CWOH-UCF101. This rate means that an only-context feature can perform well without a human in a video sequence. Thus, a context is very strong enough to cheat human action recognition, no matter what the human does in a video sequence. Figure~\ref{fig:woh_vis} shows the examples of CWOH (Figure~\ref{fig:woh001} and \ref{fig:woh002}) and MO (Figure~\ref{fig:woh003}) with ResNet-C3D. The sophisticated motion representation enables to show a lack and possibility of human action recognition.

\section{Discussion}

In this paper, we discuss the current state of motion representation, and discover new questions to investigate in the future.

\textbf{Context effects.} Can we detect human action in videos with a contextual feature ? In this study, we find the answer to be yes. Surprisingly, contextual features recorded successful rate of 53\% on the CWOH-UCF101. Using context has a great potential to aid in classifying human actions . However, the problem of recognizing human actions was swapped to become a simple scene recognition. If we look at the negative result positively , we must focus on the region around a human . For example, we should annotate contextual information such as objects and scenes in addition to human actions. Moreover, ResNet-C3D achieved a rate of over 65\% on CWOH-UCF101 with only a contextual feature. The big difference between a spatial stream and ResNet-C3D is 3D convolution. 3D convolution simultaneously processes both spatial and temporal space. Therefore, there may be a benefit from motion representation learning in a background . Although recent motion representation techniques focus on (only) human actions, we positively utilize movement of the background to train a general motion . There are several video datasets for human action recognition, such as Kinetics Human~\cite{CarreiraCVPR2017} and ActivityNet~\cite{ActivityNet}; however, no general-purpose motion dataset like ImageNet~\cite{RussakovskyIJCV2015} exists in image classification.

\textbf{Better motion representation.} Two properties, optical-flow images and 3D convolution, are important for motion-only datasets. The most impressive result on the MO dataset is given by ResNet-C3D, which gives a performance of 84\% on MO-UCF101. This result shows that 3D convolution is necessary to extract pure human motions on a human action dataset. Carreira~\textit{et al.} had outstanding performance rates on a couple of representative results~\cite{CarreiraCVPR2017}. Recently, network architecture with 3D convolution has tended to be deeper, in the manner of 2D convolution. A possible strategy is to implement a combination of deeper architecture and 3D convolution for better motion representation. One more possibility is to effectively utilize contextual features in motion representation in order to achieve a higher-level understanding. Two-stream architecture with motion and context will help in understanding high-level categories in addition to human actions.

\section{Conclusion}

In this paper, we proposed a novel evaluation of human action and context that unveils a lack of current deep neural network (DNN)-based human action recognition. We have verified the context effects of pure human motion in video-based action recognition. To execute our concept, we have created four data settings \{without-human setting  (WOH), with-human setting  (WH), complete without-human setting (CWOH), motion-only setting (MO)\} based on representative human action datasets UCF101 and HMDB51. We also provide a detailed investigation of human action recognition without a human. The effects of the context are large when using the self-created datasets. With modified UCF101, two-stream ConvNets can perform successful classification at a rate of 53.07\% with only background areas. Using the MO setting, ResNet-C3D had the best rate performance, 84.21\%, with MO-UCF101, and so a more sophisticated representation of human-centered motion should use a ConvNet-based approach. This may be a key component for improving the level of ConvNet-based motion representation in videos.

%
%
\bibliographystyle{splncs04}
\bibliography{main}

\begin{thebibliography}{10}
\providecommand{\url}[1]{\texttt{#1}}
\providecommand{\urlprefix}{URL }
\providecommand{\doi}[1]{https://doi.org/#1}

\bibitem{CarreiraCVPR2017}
Carreira, J., Zisserman, A.: Quo vadis, action recogni- tion? a new model and the kinetics dataset. IEEE Conference on Computer Vision and Pattern Recognition (CVPR) (2017)

\bibitem{EvertsCVPR2013}
Everts, I., Gernert, J.C., Gevers, T.: Evaluation of color stips for human action recognition. IEEE Conference on Computer Vision and Pattern Recognition (CVPR) (2013)

\bibitem{ActivityNet}
Heilbron, F.C., Escorcia, V., Ghanem, B., Niebles, J.C.: Activitynet: A large-scale video benchmark for human activity understanding. IEEE Conference on Computer Vision and Pattern Recognition (CVPR) (2015)

\bibitem{JainCVPR2015}
Jain, M., van Gemert, J.C., Snoek, C.G.M.: What do 15,000 object categories tell us about classifying and localizing actions? IEEE Conference on Computer Vision and Pattern Recognition (CVPR) (2015)

\bibitem{HMDB51}
Kuehne, H., Jhuang, H., Garrote, E., Poggio, T., Serre, T.: {HMDB}: a large video database for human motion recognition. International Conference on Computer Vision (ICCV) (2011)

\bibitem{LaptevIJCV2005}
Laptev, I.: On space-time interest points. International Journal of Computer Vision (IJCV) (2005)

\bibitem{LaptevICCV2003}
Laptev, I., Lindeberg, T.: Space-time interest points. International Conference of Computer Vision (ICCV) (2003)

\bibitem{LaptevCVPR2008}
Laptev, I., Marszalek, M., Schmid, C., Rozenfeld, B.: Learning realistic human actions from movies. IEEE Conference on Computer Vision and Pattern Recognition (CVPR) (2008)

\bibitem{LongCVPR2015}
Long, J., Shelhamer, E., Darrell, T.: Fully convolutional networks for semantic segmentation. IEEE Conference on Computer Vision and Pattern Recognition (CVPR) (2015)

\bibitem{MarszalekCVPR2009}
Marszalek, M., Laptev, I., Schmid, C.: Actions in context. IEEE Conference on Computer Vision and Pattern Recognition (CVPR) (2009)

\bibitem{RenNIPS2015}
Ren, S., He, K., Girshick, R., Sun, J.: Faster r-cnn: Towards real-time object detection with region proposal networks. Neural Information Processing Systems (NIPS) (2015)

\bibitem{RussakovskyIJCV2015}
Russakovsky, O., Deng, J., Su, H., Krause, J., Satheesh, S., Ma, S., Huang, Z., Karpathy, A., Khosla, A., Bernstein, M., Berg, A.C., Fei-Fei, L.: Imagenet large scale visual recognition challenge. International Journal of Computer Vision (IJCV) (2015)

\bibitem{SimonyanNIPS2014}
Simonyan, K., Zisserman, A.: Two-stream convolutional networks for action recognition. Neural Information Processing Systems (NIPS) (2014)

\bibitem{UCF101}
Soomro, K., Zamir, A.R., Shah, M.: Ucf101: A dataset of 101 human action classes from videos in the wild. CRCV-TR-12-01 (2012)

\bibitem{TranICCV2015}
Tran, D., Bourdev, L., Fergus, R., Torresani, L., Paluri, M.: Learning spatiotemporal features with 3d convolutional networks. International Conference on Computer Vision (ICCV) (2015)

\bibitem{TranarXiv2017}
Tran, D., Ray, J., Shou, Z., Chang, S.F., Paluri, M.: Convnet architecture search for spatiotemporal feature learning. arXiv pre-print: 1708.05038 (2017)

\bibitem{WangCVPR2011}
Wang, H., Klaser, A., Schmid, C., Cheng-Lin, L.: Action recognition by dense trajectories. IEEE Conference on Computer Vision and Pattern Recognition (CVPR) (2011)

\bibitem{WangICCV2013}
Wang, H., Schmid, C.: Action recognition with improved trajectories. International Conference on Computer Vision (ICCV) (2013)

\bibitem{WangBMVC2009}
Wang, H., Ullah, M.M., Klaser, A., Laptev, I., Schmid, C.: Evaluation of local spatio-temporal features for action recognition. British Machine Vision Conference (BMVC) (2009)

\bibitem{WangCVPR2015}
Wang, L., Qiao, Y., Tang, X.: Action recognition with trajectory-pooled deep-convolutional descriptors. IEEE Conference on Computer Vision and Pattern Recognition (CVPR) (2015)

\bibitem{LiminWangIJCV2017}
Wang, L., Wang, Z., Qiao, Y., Gool, L.V.: Transferring deep object and scene representations for event recognition in still images. International Journal of Computer Vision (IJCV) (2017)

\bibitem{WangarXiv2015}
Wang, L., Xiong, Y., Wang, Z., Qiao, Y.: Towards good practices for very deep two-stream convnets. arXiv pre-print 1507.02159 (2015)

\bibitem{WangECCV2016}
Wang, L., Xiong, Y., Wang, Z., Qiao, Y., Lin, D., Tang, X., Gool, L.V.: Temporal segment networks: Towards good practices for deep action recognition. European Conference on Computer Vision (ECCV) (2016)

\bibitem{WuCVPR2016}
Wu, Z., Fu, Y., Jiang, Y.G., Sigal, L.: Harnessing object and scene semantics for large-scale video understanding. IEEE Conference on Computer Vision and Pattern Recognition (CVPR) (2016)

\bibitem{Places205}
Zhou, B., Lapedriza, A., Xiao, J., Torralba, A., Oliva, A.: Learning deep features for scene recognition using places database. Advances in Neural Information Processing Systems (NIPS) (2014)

\end{thebibliography}
\end{document}